%% file: VLcontrol.tex
\documentclass[conference]{IEEEtran}
\usepackage{cite}
\usepackage{amsmath,amssymb,amsfonts}
\usepackage{algorithmic}
\usepackage{graphicx}
\usepackage{textcomp}
\usepackage{xcolor}
\usepackage{booktabs}
\usepackage{multirow}
\usepackage{url}
\def\BibTeX{{\rm B\kern-.05em{\sc i\kern-.025em b}\kern-.08em
    T\kern-.1667em\lower.7ex\hbox{E}\kern-.125emX}}

\begin{document}

\title{Visual Grounding in Zero-Shot Vision--Language Control}

\author{
	\IEEEauthorblockN{J. de Curtò\IEEEauthorrefmark{1}\IEEEauthorrefmark{2}\IEEEauthorrefmark{6}, Dayani Plasencia\IEEEauthorrefmark{3}\IEEEauthorrefmark{2},  Diego S\'anchez\IEEEauthorrefmark{4}\IEEEauthorrefmark{2},  I. de Zarzà\IEEEauthorrefmark{5}\IEEEauthorrefmark{6}}
	\IEEEauthorblockA{\IEEEauthorrefmark{1}\textit{Department of Computer Applications in Science \& Engineering}, 
	\textit{BARCELONA Supercomputing Center}, 
	Barcelona, Spain}
\IEEEauthorblockA{\IEEEauthorrefmark{2}\textit{Escuela Técnica Superior de Ingeniería (ICAI)}, 
\textit{Universidad Pontificia Comillas}, 
Madrid, Spain \\
Email: jdecurto@icai.comillas.edu}
\IEEEauthorblockA{\IEEEauthorrefmark{3}\textit{Herbert Wertheim College of Engineering}, 
	\textit{University of Florida}, 
	Gainesville, FL 32611, USA\\
	Email: dayani.plasencia@ufl.edu}
\IEEEauthorblockA{\IEEEauthorrefmark{4}\textit{The Grainger College of Engineering}, 
	\textit{University of Illinois Urbana-Champaign}, 
	Urbana, IL 61801, USA\\
	Email: dsanc83@illinois.edu}
\IEEEauthorblockA{\IEEEauthorrefmark{5}\textit{Human centered AI, Data \& Software}, 
	\textit{LUXEMBOURG Institute of Science and Technology}, 
	Esch-sur-Alzette, Luxembourg \\
	Email: irene.zarza@list.lu}		
\IEEEauthorblockA{\IEEEauthorrefmark{6}\textit{Estudis d'Informàtica, Multimèdia i Telecomunicació}, 
	\textit{Universitat Oberta de Catalunya}, 
	Barcelona, Spain}	
}

\maketitle

\begin{abstract}
Vision--language models (VLMs) are increasingly used as zero-shot controllers,
but successful trajectories do not necessarily show that their decisions are
grounded in visual input: simulator dynamics and conservative action priors can
produce favourable scores without meaningful perception. We investigate this
problem with an input-ablation battery combining blind-image controls, repeated
identical inputs, lane-axis reflection, non-visual baselines, and
pipeline-integrity checks. Across nine direct-action models, six structured
local VLMs, and an exploratory VLM--MPC hierarchy, we analyse $32\,874$ scored
calls over two embodiments and three simulation environments. The direct-control
results are largely negative. A constant-\textsc{slow} policy outperforms a
scripted geometric controller, several models are image-invariant or nearly
constant, and models that recognize longitudinal hazards still fail to transform
\textsc{left} and \textsc{right} decisions consistently under reflection. None
of the six local VLMs satisfies the joint longitudinal and lateral grounding
criteria. However, an image-only deterministic positive control estimates the
lead gap with $0.090$~m MAE and exact mirror equivariance, confirming that the
stimuli and controller interface contain sufficient visual information. The failures are therefore modular rather than universal. A post-hoc but
leakage-controlled symmetry-consensus guardian selects Gemma4-12B and
Qwen3.5-9B using 16 calibration frames and freezes a 2-of-4 hazard vote across
original and reflected views. On 272 held-out frames, it achieves $0.956$
accuracy and $0.954$ balanced accuracy (episode-cluster bootstrap 95\% CI
$[0.895,0.990]$). Nested leave-one-episode-out evaluation recovers the same
model pair and threshold in all 12 folds and reaches $0.951$ balanced accuracy.
Abstaining on tied votes raises committed balanced accuracy to $0.973$ at
$0.824$ coverage. When deterministic perception retains lateral authority,
offline modular replay achieves $0.934$ action agreement and exact mirror
equivariance. A separate MPC diagnostic further shows that low-level
optimization cannot recover missing geometry from collapsed visual intent.
These results support the use of current VLMs as bounded, selective hazard
assistants rather than as monolithic zero-shot controllers.
\end{abstract}


%% =====================================================================
\section{Introduction}

Vision--language models have made zero-shot visual control an appealing
proposition for networked autonomous systems. A single pretrained model,
prompted in natural language, promises to replace a hand-engineered perception
and planning stack, and the same model can be served from an on-board
accelerator or from a hosted endpoint reached over the network. Both
deployments are now practical, and the choice between them is a systems
question involving latency, energy and recurring cost.

That systems question presupposes a prior one that is rarely asked: does the
controller use the camera at all?

The concern is concrete. Simulated driving benchmarks reward forward progress
and penalise collisions. In such environments a policy that always decelerates
is close to optimal, and it requires no perception. If a VLM controller emits a
near-constant action, aggregate metrics will report competence that the model
does not have, and the resulting deployment decision, edge or cloud, which
model, at what cost, will be made on evidence that measures the simulator
rather than the model.

We make both the failure and its useful boundary measurable:

\begin{enumerate}
\item an input-ablation battery with blind controls, a noise floor,
lane-axis reflection, baseline policies, and four pipeline-integrity guards
(Sections~\ref{sn:battery});
\item a constructive bounded capability: an image-only positive control
and a leakage-controlled symmetry-consensus guardian reaching $0.954$ held-out
balanced accuracy and $0.973$ at $0.824$ selective coverage
(Sections~\ref{sn:constructive} and~\ref{sn:guardian-results});
\item evidence that monolithic scores and downstream sophistication do
not certify perception: constants can outperform scripts, and visual intent
reduces an MPC hierarchy from $23/25$ to $8/25$ goals; and
\item a graded capability structure in which longitudinal hazard
recognition can be recovered despite failed lateral geometry, while prompt
formulation strongly changes measured grounding
(Sections~\ref{sn:threshold} and~\ref{sn:prompts}).
\end{enumerate}

%% =====================================================================
\section{Background and Related Work}

Vision--language pretraining \cite{radford2021clip,liu2023llava} supports
open-ended image reasoning and has been extended to direct action selection
\cite{zhou2024navgpt,song2024vlm,elnoor2025vlm,zhang2025mapnav}, driving-hazard
description \cite{xiao2024hazardvlm}, and mixed aerial--ground perception
\cite{LLMfusiondeCurto2025,deCurto2023}. Open-weight controllers range from
compact edge models \cite{marafioti2025smolvlm,yao2024minicpmv} to hosted 72B
systems \cite{wang2024qwen2vl,bai2025qwen25vl}.

Shortcut performance can be unrelated to the intended capability
\cite{geirhos2020shortcut}, motivating behavioural tests
\cite{ribeiro2020beyond,liang2022holistic}. This risk is acute in control,
where dynamics, safety logic and reward shape trajectories independently of the
policy. We therefore combine input ablation with metamorphic relations
\cite{chen2018metamorphic}, an approach used in autonomous-driving perception
\cite{tian2018deeptest,zhang2018deeproad} and semantic-invariance testing
\cite{metamorphic2025deCurto,deZarza2026_2}, and report effect sizes rather than
aggregate success alone.

On-board inference bounds latency and keeps data local but consumes limited
memory and energy \cite{marafioti2025smolvlm,yao2024minicpmv,zhang2024edgeshard};
hosted inference offers larger models at the cost of network delay, recurring
charges and less control over the stack. We compare both on identical stimuli
(Section~\ref{sn:cost}). Local energy is directly observable
\cite{luccioni2024power}; hosted energy is not derivable from the API.

%% =====================================================================
\section{The Input-Ablation Battery}\label{sn:battery}

The battery asks whether the action depends on the image and, if so, on which
property, using tests of increasing specificity.

For each frame we issue the same prompt under eight input conditions
(Table~\ref{t:conditions}). Only the image varies. Agreement is measured
against the action produced for the real frame.

\begin{table}[t]
\caption{Blind-input conditions. Only the image changes.}
\label{t:conditions}
\centering
\scriptsize
\begin{tabular}{@{}lll@{}}
\toprule
\textbf{Condition} & \textbf{Image} & \textbf{Tests} \\
\midrule
\texttt{real}            & true frame               & reference \\
\texttt{real\_repeat}    & true frame, 2nd call     & noise floor \\
\texttt{blank}           & uniform mid-grey         & is the image read at all? \\
\texttt{noise}           & uniform random RGB       & is any content used? \\
\texttt{shuffled}        & other frame, same scen.  & is \emph{this} frame used? \\
\texttt{cross\_scenario} & frame, other scenario    & is scene context used? \\
\texttt{text\_only}      & none                     & prompt alone sufficient? \\
\texttt{mirrored}        & flip across lane axis    & lateral grounding \\
\bottomrule
\end{tabular}
\end{table}

We report $\mathrm{IDI}=1-\mathrm{agree}(\texttt{real},\texttt{blank})$,
but use a condition--action $\chi^2$ test with Cram\'er's $V$ as the primary,
Holm-corrected statistic. IDI is weak: a pixel-grounded synthetic policy scored
only $0.083$ on mostly symmetric frames, whereas $V=0.61$ detected dependence.

Hosted endpoints remain stochastic at temperature $0$. The
\texttt{real\_repeat} condition measures self-disagreement $s$; dependence below
that floor is uninformative. We report
$\mathrm{IDI}_{\mathrm{adj}}=\mathrm{IDI}-s$ and flag IDI $\leq1.5s$. One model
repeated its answer on only $21\%$ of identical images ($s=0.79$): apparent
IDI was $0.77$, but adjusted IDI was $-0.021$, despite conventional tests giving
$p<10^{-4}$.

Refusal or unparseable blank-image output changes the response, not necessarily
the decision. We test both all outputs and parsed actions; significance only at
response level is a refusal effect. One model had $V=0.46$, $p<10^{-4}$ over
outputs but $V=0$, $p=1$ over decisions, entirely from $54/144$ blank-frame
parse failures versus $1/144$ real-frame failures.

Under reflection across the lane axis, grounded direction must exchange
\textsc{left} and \textsc{right}; failure to swap indicates a prior. For this
top-down horizontal road, the correct image transform is a vertical flip, not a
travel-reversing horizontal flip. We report $n_{\mathrm{dir}}$ because the test
has no power without directional actions.

We test real-frame action against lead gap $g$ using both a contingency test for
$g<18$\,m and permutation-tested $I(A;g)$ with quartile-binned $g$. Grounding
requires both to reject and a smaller median gap for braking than proceeding.

On identical seeds we compare every constant action, uniform random control,
sampling from each model's action marginal, and a scripted geometric controller
using ground-truth state.

Because steps are autocorrelated, confidence intervals cluster-bootstrap episodes
($B=5000$); permutation tests also use 5000 resamples. Model comparisons are
Holm-corrected \cite{holm1979simple} using SciPy \cite{virtanen2020scipy}.

%% =====================================================================
\section{Experimental Setup}

We use two embodiments: a quadrotor in \texttt{gym-pybullet-drones}
\cite{panerati2021gympybullet} and a ground vehicle in \texttt{highway-env}
\cite{leurent2018highwayenv}, both under Gymnasium
\cite{towers2023gymnasium}. The blind-input battery uses a static frame bank of
144 frames (laptop arm) and 288 frames (GPU arm) drawn from four scenarios,
light traffic, dense traffic, a merge, and a blocked lane, at three seeds
each. Because rendering differs across operating systems and SDL builds while
simulation does not, we verify cross-machine equivalence by a \emph{geometry
fingerprint}: a hash of every frame's ground-truth lead gap. The two arms in
this work agree exactly, on all $144$ shared frames
to $0.000$\,m, despite different \texttt{pygame} and \texttt{numpy} versions.

\subsection{Models}
The edge arm comprises four open-weight models runnable on a single T4:
SmolVLM-Instruct \cite{marafioti2025smolvlm}, Qwen2-VL-2B and Qwen2-VL-7B
\cite{wang2024qwen2vl}, and LLaVA-1.5-7B \cite{liu2023llava}, loaded through the
Transformers library \cite{wolf2020transformers}. The hosted arm comprises
Qwen2.5-VL-72B \cite{bai2025qwen25vl}, MiniCPM-V-4.5
\cite{yao2024minicpmv,yu2025minicpmv45}, Cosmos3-Super-Reasoner (33B), Kimi-K2.6
and Kimi-K3, served through an OpenAI-compatible endpoint.

The structured local arm uses Qwen3-VL-8B \cite{bai2025qwen3vl}, Gemma~4-12B
\cite{gemmateam2026gemma4}, Qwen3.5-9B \cite{qwen2026qwen35}, Qwen2.5-VL-7B,
Ministral~3-8B \cite{mistral2025mistral3}, and MiniCPM-V-4.5: recorded Q4\_K\_M
Ollama~0.32.5 builds run sequentially on one A100-80GB. Preflight pins one
request mode per model; exact digests and templates are released.

Reasoning models emit a chain of thought before the answer; if the token budget
is exhausted first, the visible content is empty and the call is lost. We
therefore set the budget per model from a measured probe rather than one global
value, and request structured output where the endpoint supports it. Under a
shared $1024$-token budget the two Kimi models lost $64\%$ and $60\%$ of calls to
truncation, while Cosmos3 completed in a mean of $187$ tokens; at $4096$ tokens
with JSON-mode output those losses fall to $28\%$ and $0.7\%$. Every truncated
call is recorded as a failure and never silently substituted.

\subsection{Corpus and parsing}
The original study contributes $23\,416$ model queries: $2\,016$ edge, two
$5\,760$-query hosted runs, $9\,000$ prompt-ablation calls and $880$ probes. The
structured local arm adds $192$ admission, $3\,456$ static original/mirror and
$1\,585$ online queries. The MPC diagnostic adds $4\,225$ Gemma4 calls across
$531$ episodes, for $32\,874$ scored calls excluding adapter preflight.
Ground-truth state is never supplied to a model. Parse failures remain
\texttt{UNPARSED}; online fail-safe execution is logged separately as
\texttt{executed\_action=\textsc{slow}} and is never credited to the model.

\subsection{Constructive renderer-calibrated controller}
\label{sn:constructive}
Structured scene estimates feed a fixed safety policy. The image-only positive
control measures geometry from RGB without simulator state; VLMs receive a
static crop/enlargement with a renderer legend, 10~m ruler and fixed thresholds,
but no frame-specific label. Admission on 16 balanced frame/mirror pairs
requires parse $\geq.95$, raw and balanced close accuracy $\geq.75$, scene/action
equivariance $\geq.65/.75$, at least three actions and two directional actions.
The 288-pair full battery and 12-episode online diagnostic retain the same gate;
online frames additionally pass renderer-integrity and positive-control checks.

\begin{figure*}[t]
    \centering
    \includegraphics[width=\textwidth]{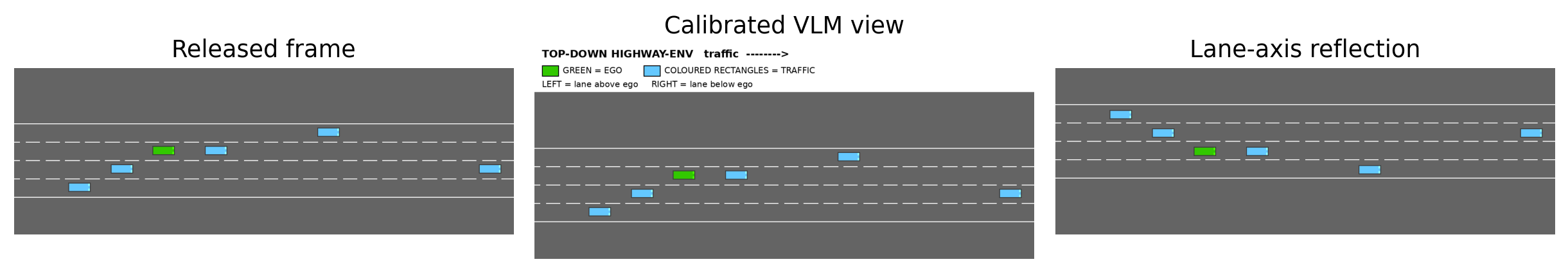}
    \caption{Visual interface used by the structured controller. From left to right: the released \texttt{highway-env} frame, the calibrated VLM view with an explicit identification of the ego vehicle and left--right lane convention, and the reflection across the lane axis used to evaluate symmetry consistency. Under this transformation, longitudinal hazard estimates should remain unchanged, whereas \textsc{Left} and \textsc{Right} decisions should be exchanged.}
    \label{fgr:vlm-interface}
\end{figure*}

The structured controller operates on a calibrated top-down representation with an explicit ego reference and lane convention (see Figure~\ref{fgr:vlm-interface}). In addition to the released simulator frame, we provide a legend-augmented view and evaluate symmetry consistency by reflecting the scene across the lane axis. 

\subsection{Symmetry-consensus longitudinal guardian}
\label{sn:guardian-method}
We reuse the completed local calls in a post-hoc but leakage-controlled
analysis. The pre-existing 16-frame admission set selects two of six models and
a threshold over four hazard votes (two models $\times$ original/reflected
view); both are frozen before the other 272 frames are read. A 2--2 split permits
abstention. Episode-cluster bootstrap and nested leave-one-episode-out tests
measure uncertainty and selection stability. In offline modular replay, the
VLMs receive only longitudinal authority while a deterministic RGB parser
retains collision and lateral-clearance authority.

\subsection{Hierarchical VLM--MPC diagnostic}
A separate forward-camera rover tests whether a competent low-level controller
can rescue a weak semantic planner. The default Gemma4 Ollama interface emits
one of five intents every $1.0$~s; a constrained, linearized bicycle MPC
\cite{mayne2000mpc} runs at $0.2$~s with horizon eight and maps the intent to a
target speed and lateral offset. In the primary $5$-scenario $\times$ $5$-seed
ablation, the MPC is fixed while the intent source varies and simulator obstacle
centres feed its keep-out constraints. A second $4$-scenario $\times$ $8$-seed
ablation disables those constraints except for an oracle-MPC reference and
separates lateral-only from full VLM directives. Safety overrides are disabled;
the episode records report zero unparsed outputs across $4\,225$ VLM calls. The
exact model digest and the separate ablation driver were not present in the returned bundle, so this arm is explicitly
exploratory; controller source, episode records and an independent validator are
released.

%% =====================================================================
\section{Results}

Table~\ref{t:baselines} reports the baseline ladder over
$7$ policies $\times\ 4$ scenarios $\times\ 8$ seeds.

\begin{table}[t]
\caption{Closed-loop baseline ladder (\texttt{highway-env}), 32 episodes per
policy. A constant action outperforms the scripted controller on every metric.}
\label{t:baselines}
\centering
\small
\begin{tabular}{@{}lrrrr@{}}
\toprule
\textbf{Policy} & \textbf{Dist.\,(m)} & \textbf{Reward} & \textbf{Speed} & \textbf{Crash} \\
\midrule
\textbf{constant-\textsc{slow}} & \textbf{514.4} & \textbf{19.64} & 20.1 & \textbf{3.1\%} \\
oracle-geometric      & 444.5 & 16.34 & 23.5 & 37.5\% \\
random-uniform        & 339.3 & 12.05 & 22.7 & 78.1\% \\
constant-\textsc{left}  & 211.3 & 6.92 & 25.0 & 87.5\% \\
constant-\textsc{right} & 197.0 & 7.01 & 25.0 & 100\% \\
constant-\textsc{straight} & 145.8 & 5.13 & 28.1 & 100\% \\
\bottomrule
\end{tabular}
\end{table}

The best policy is a constant one. It travels $15.7\%$ farther than the scripted
geometric controller, earns $20.2\%$ more reward, and crashes $12\times$ less
often, while consuming no visual input whatsoever. Any VLM that converges on
deceleration will therefore score well on aggregate metrics, and three of the
models we test do exactly that.

Two caveats. \texttt{highway-env}'s discrete meta-action space has no true stop,
so \textsc{stop} and \textsc{slow} alias to the same command and produce
identical rows; we report one. And the scripted controller's $37.5\%$ crash rate
shows it is a weak reference, not an upper bound: its \textsc{straight} default
at high speed causes collisions. The comparison that matters is against the
constants, which require no tuning.

A decision-level replay reaches the same conclusion: constant-\textsc{straight}
matches or exceeds every edge model's agreement with the scripted controller,
and randomizing the temporal order while preserving each model's action
histogram does not reduce two of the four scores. We release the complete replay
table rather than treating agreement with a weak oracle as a principal result.

\subsection{Structured output is not geometric grounding}

All six VLMs pass image-path, schema and online-renderer preflights, use a
pinned request mode, parse every request and require no fallback. The positive
control recovers 274 visible lead gaps to $0.090$~m MAE with exact scene/action
equivariance and all five actions (Table~\ref{t:controller-local}); the VLM
null result is therefore not a pipeline or stimulus failure.

\begin{table}[t]
\caption{Task-calibrated local arm: 288 static original/mirror pairs and 12
valid-renderer episodes per estimator. Swap denominators are in parentheses; no
VLM passes the joint gate.}
\label{t:controller-local}
\centering
\scriptsize
\setlength{\tabcolsep}{1.6pt}
\resizebox{\columnwidth}{!}{\input{tables/controller_local_vlms.tex}}
\end{table}

Gemma~4 and Qwen3.5 classify proximity well (balanced accuracy $0.858$ and
$0.867$) but have scene equivariance $0.399$ and $0.031$. Ministral~3 has the
lowest MAE ($6.240$~m) by over-calling danger (sensitivity/specificity
$0.995/0.074$). Qwen2.5-VL's action equivariance $0.872$ is degenerate: only one
directional decision is evaluable and scene equivariance is zero. The best
correct directional swap is $0.134$; none is admitted.

Yet Qwen3-VL, Qwen3.5, Ministral~3 and MiniCPM each crash in only $1/12$
episodes. Qwen3.5 emits \textsc{slow} on $308/310$ decisions. These trajectories
reflect a conservative prior and simulator reward, not reliable geometry.

\subsection{Symmetry consensus recovers a longitudinal guardian}
\label{sn:guardian-results}
Failure of the five-action gate does not remove every useful visual signal.
Calibration selects Gemma~4, Qwen3.5 and a 2-of-4 threshold. On the frozen
272-frame holdout, original/reflected consensus obtains $0.956$ accuracy and
$0.954$ balanced accuracy, with sensitivity/specificity $0.958/0.950$
(Table~\ref{t:positive-guardian}) and a cluster-bootstrap 95\% interval
$[0.895,0.990]$. It corrects 35 Gemma~4 errors while introducing four (exact
McNemar $p=3.35\times10^{-7}$).

\input{tables/positive_guardian.tex}

Nested leave-one-episode-out training selects the same pair and threshold in
$12/12$ folds and reaches $0.951$ balanced accuracy. Delegating 2--2 splits to a
classical controller raises committed balanced accuracy to $0.973$ at $0.824$
coverage; unanimity reaches $0.996$ at $0.647$. Offline modular replay retains
deterministic lateral geometry, agrees with the renderer policy on $0.934$ of
holdout frames and is exactly mirror-equivariant. Only seven reference actions
are directional, so the positive claim is deliberately narrow: VLMs can serve
as a selective longitudinal hazard module, not a learned lateral controller.

\subsection{MPC cannot repair an ungrounded intent layer}
\label{sn:mpc}
The two-rate hierarchy improves low-level tracking but not the missing visual
semantics. In $25$ paired worlds, changing only the high-level source yields
$23/25$ goals for MPC-only and oracle intent, $14/25$ for random intent, and
$8/25$ for both the VLM and its mirrored-input condition
(Table~\ref{t:mpc-intent}). The VLM emits \textsc{stop} on $81.6\%$ of steps;
mirroring changes neither completion nor the absence of \textsc{right} actions.
Blank input produces \textsc{stop} on all $5\,000$ steps. A trivial
constant-\textsc{left} source reaches $24/25$, again showing that completion is
not a grounding metric. The zero collision rate is not visual safety: this
primary MPC receives simulator obstacle centres.

\begin{table}[t]
\caption{Selected intent-source controls with the state-constrained MPC fixed,
$25$ paired worlds per source. The full 11-source table is released.}
\label{t:mpc-intent}
\centering
\scriptsize
\setlength{\tabcolsep}{2.5pt}
\input{tables/mpc_intent_ablation.tex}
\end{table}

The harder ablation exposes that dependence. Constrained \texttt{mpc-oracle}
reaches $30/32$ goals with $2/32$ collisions; \texttt{mpc-blind} falls to $1/32$
and $31/32$. Lateral VLM intent reaches $0/32$ with $32/32$ collisions, while
its mirror reaches $5/32$ with $27/32$. Full VLM intent reaches $0/32$ but cuts
collisions to $11/32$ by issuing \textsc{stop} on $3\,970/4\,554$ steps. MPC can
execute valid geometry; it cannot create it from collapsed intent.

\subsection{Edge models are image-invariant}

Table~\ref{t:degeneracy} audits the closed-loop logs. Five of eight
model--platform cells are constant policies (modal share $\geq 0.98$). Three
show scene-dependence surviving Holm correction, but at magnitudes that do not
support a claim of situational awareness: the largest, SmolVLM on the aerial
platform, explains $8.0\%$ of its own action variability, and the rover cells
explain $3.5\%$ and $28.3\%$ of a near-zero entropy.

\begin{table}[t]
\caption{Policy degeneracy audit on closed-loop logs. $\mathrm{MI}/H$ is the
fraction of the model's own action variability explained by scene geometry.
$p$ is Holm-corrected.}
\label{t:degeneracy}
\centering
\scriptsize
\begin{tabular}{@{}llrrrrr@{}}
\toprule
\textbf{Plat.} & \textbf{Model} & \textbf{Modal} & \textbf{$H$\,(b)} &
\textbf{MI\,(b)} & \textbf{MI/$H$} & \textbf{$p$} \\
\midrule
drone & Qwen2-VL-2B  & 1.000 & 0.00 & --- & --- & --- \\
drone & Qwen2-VL-7B  & 1.000 & 0.00 & --- & --- & --- \\
drone & LLaVA-1.5-7B & 0.928 & 0.45 & 0.032 & 0.071 & 0.373 \\
drone & SmolVLM      & 0.485 & 1.82 & 0.146 & 0.080 & \textbf{0.002} \\
rover & Qwen2-VL-2B  & 1.000 & 0.00 & --- & --- & --- \\
rover & LLaVA-1.5-7B & 1.000 & 0.00 & --- & --- & --- \\
rover & Qwen2-VL-7B  & 0.982 & 0.13 & 0.037 & 0.283 & \textbf{0.037} \\
rover & SmolVLM      & 0.649 & 0.94 & 0.033 & 0.035 & \textbf{0.041} \\
\bottomrule
\end{tabular}
\end{table}

Two models produce \emph{numerically identical} trajectories on $10$
scenario--seed cells: both are constant-\textsc{straight} policies in a
deterministic simulator, which is a genuine convergence rather than a logging
error, but one that would be indistinguishable from a caching bug without the
degeneracy audit. Qwen2-VL-2B additionally fails to parse on $54.4\%$ of aerial
steps, a fact invisible in any aggregate success rate.

\begin{figure}[t]
\centering
\includegraphics[width=\columnwidth]{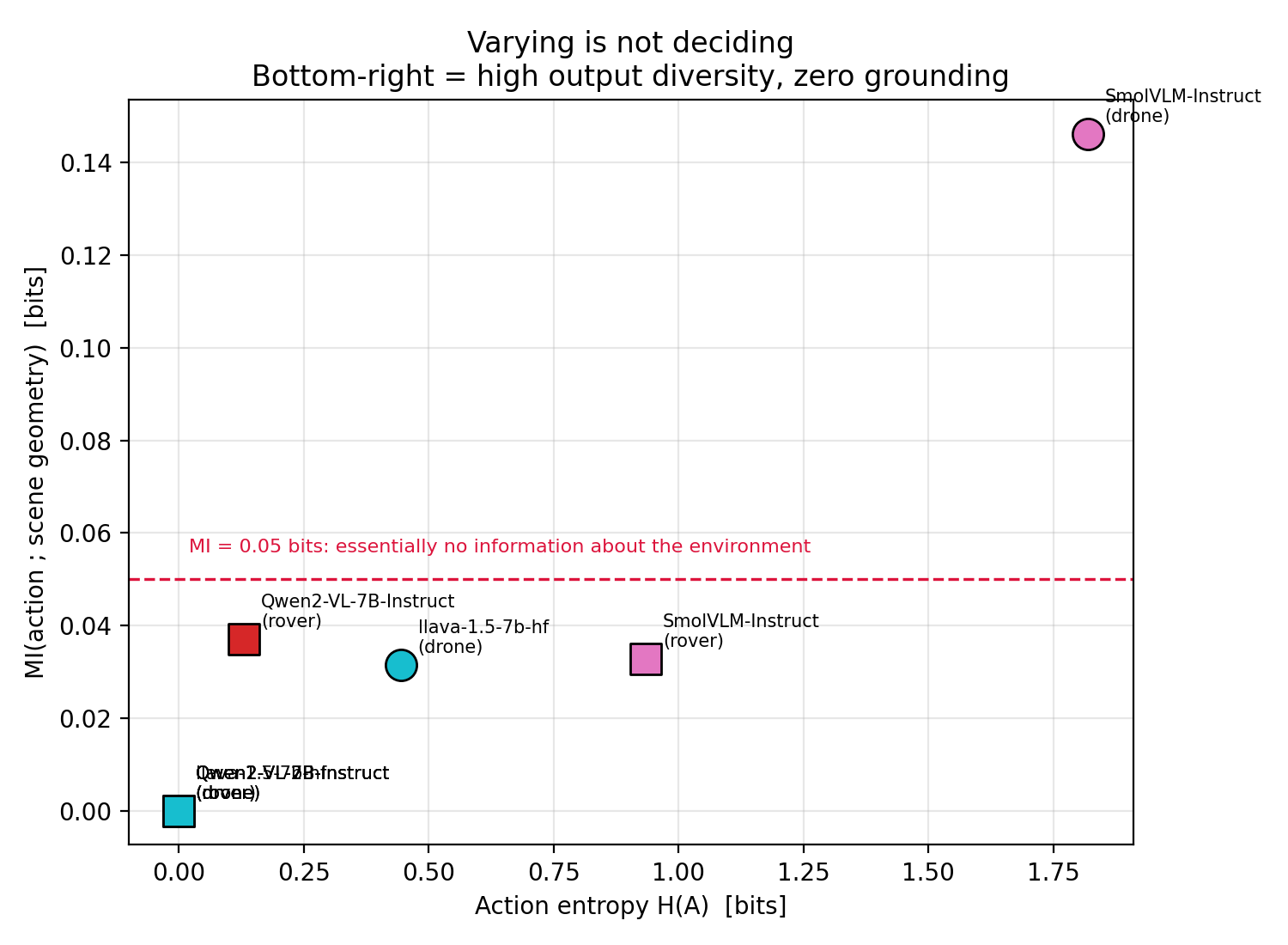}
\caption{Action entropy against mutual information with scene geometry. High
entropy with near-zero MI (lower right) is the diagnostic pathology: output
diversity without environmental grounding.}
\label{fgr:mi}
\end{figure}

Figure~\ref{fgr:mi} separates varying from deciding. SmolVLM has the highest
action entropy of any edge model and among the lowest information content: it
fluctuates without tracking the world.

The blind-input battery corroborates this on the one edge model whose vision
path passed our loader sanity check: SmolVLM shows
$\mathrm{agree}(\texttt{blank})=0.903$ but
$\mathrm{agree}(\texttt{text\_only})=0.097$. Replacing the road with a grey
rectangle barely changes its action; removing the image entirely changes it on
$90\%$ of frames. The \emph{presence} of an image token matters far more than
its content.

\subsection{Four regimes in the hosted arm}
\label{sn:threshold}

Table~\ref{t:frontier} reports the five hosted models. Four regimes are
visible, and the fourth is only visible because of the noise floor.

\begin{table}[t]
\caption{Hosted arm, 144 frames $\times$ 8 conditions per model. $s$ is the
self-consistency noise floor: the rate at which a model disagrees with itself on
the \emph{identical} image. Values in the last two rows are computed on parsed
decisions only.}
\label{t:frontier}
\centering
\scriptsize
\setlength{\tabcolsep}{2pt}
\begin{tabular}{@{}lrrrrr@{}}
\toprule
& \textbf{Qwen2.5} & \textbf{Cosmos3} & \textbf{MiniCPM} & \multicolumn{2}{c}{\textbf{Kimi}} \\
\cmidrule(lr){2-2}\cmidrule(lr){3-3}\cmidrule(lr){4-4}\cmidrule(lr){5-6}
\textbf{Metric} & \textbf{VL-72B} & \textbf{SR-33B} & \textbf{V-4.5} & \textbf{K2.6} & \textbf{K3} \\
\midrule
unparsed rate             & 0.000 & 0.000 & 0.000 & 0.282 & 0.007 \\
noise floor $s$           & \textbf{0.021} & 0.125 & \textbf{0.000} & 0.410 & \textbf{0.792} \\
agree(\texttt{blank})     & 0.396 & 0.222 & 0.000 & 0.007 & 0.229 \\
agree(\texttt{noise})     & 0.396 & 0.090 & 0.000 & 0.000 & 0.188 \\
agree(\texttt{text\_only})& 0.396 & 0.222 & 0.000 & 0.000 & 0.167 \\
agree(\texttt{cross\_sc.})& \textbf{0.333} & 0.479 & 0.986 & 0.424 & 0.208 \\
Cram\'er's $V$            & 0.502 & 0.557 & 0.710 & 0.453 & 0.302 \\
$\mathrm{IDI}_{\mathrm{adj}}$ & 0.583 & 0.653 & 1.000 & 0.583 & \textbf{$-0.021$} \\
clears noise floor        & yes & yes & yes & yes & \textbf{no} \\
$n_{\mathrm{dir}}$        & 50 & 13 & 0 & 3 & 40 \\
mirror swap rate          & \textbf{0.000} & \textbf{0.000} & --- & 0.000 & 0.225 \\
$I(A;g)$ (bits)           & \textbf{0.408} & \textbf{0.198} & 0.028 & 0.106 & 0.122 \\
$p$ (permutation)         & \textbf{0.0002} & \textbf{0.0002} & 0.125 & 0.244 & 0.030 \\
$p$ (action vs.\ obs.)& \textbf{$<10^{-4}$} & \textbf{0.0016} & 0.370 & 0.483 & 0.120 \\
\midrule
\textbf{Verdict} & grounded & grounded & presence & not & self- \\
                 & (depth)  & (depth)  & only     & scored & inconsistent \\
\bottomrule
\end{tabular}
\end{table}

\textbf{Qwen2.5-VL-72B is grounded in depth.} Its three emitted actions order
monotonically by ground-truth lead gap: \textsc{stop} at a median $5.0$\,m
($n{=}57$), \textsc{slow} at $26.8$\,m ($n{=}37$), \textsc{left} at $33.2$\,m
($n{=}50$). Mutual information with the gap is $0.408$ bits, permutation
$p=0.0002$, against a noise floor of $0.021$. Its behaviour under degraded input
is unambiguous: \texttt{blank}, \texttt{noise} and \texttt{text\_only} each
produce \textsc{stop} on all $144$ frames, $432/432$ identical responses to
three entirely different non-inputs. It stops when it cannot see and grades its
response by distance when it can. Notably, \texttt{cross\_scenario} yields the
\emph{lowest} agreement of any condition ($0.333$): a different real scene
changes its action more than no scene at all.

\textbf{Cosmos3-Super-Reasoner is grounded more coarsely.} It is the only model
to use the full five-action vocabulary, and it is also the cheapest we tested.
Its actions separate near from far ($I(A;g)=0.198$ bits, $p=0.0002$; action
against the binary obstacle event $p=0.0016$), but the ordering is not monotone:
\textsc{straight} at $22.1$\,m and \textsc{slow} at $12.3$\,m are distinguishable,
while \textsc{stop}, \textsc{left} and \textsc{right} all sit at a median $5.0$\,m,
and it still emits \textsc{straight} on $39$ of the $81$ frames where an obstacle
lies inside $18$\,m. One detail separates it from Qwen: it answers \textsc{slow}
to a blank frame and \textsc{stop} to a noise frame, whereas Qwen answers
\textsc{stop} to both. Discriminating uniform grey from random pixels is
low-level image sensitivity, and it is evidence that the model reads the image
without reading the scene.

\begin{table}[t]
\caption{The two grounded models: emitted action against the simulator's
ground-truth lead gap, real frames only. Neither ordering was prompted. Under
lane-axis reflection neither model exchanges direction.}
\label{t:qwenorder}
\centering
\scriptsize
\setlength{\tabcolsep}{3pt}
\begin{tabular}{@{}llrrl@{}}
\toprule
\textbf{Model} & \textbf{Action} & \textbf{$n$} & \textbf{Med.\ gap} & \textbf{Under reflection} \\
\midrule
\multirow{3}{*}{Qwen2.5}
 & \textsc{left}  & 50 & 33.2\,m & \textsc{slow} 25, \textsc{left} 24, \\
 & \textsc{slow}  & 37 & 26.8\,m & \textsc{stop} 1 \\
 & \textsc{stop}  & 57 & \phantom{0}5.0\,m & --- \\
\midrule
\multirow{5}{*}{Cosmos3}
 & \textsc{straight} & 86 & 22.1\,m & --- \\
 & \textsc{slow}     & 32 & 12.3\,m & --- \\
 & \textsc{stop}     & 13 & \phantom{0}5.0\,m & --- \\
 & \textsc{right}    & \phantom{0}8 & \phantom{0}5.0\,m & \textsc{straight} 6, \textsc{slow} 3, \\
 & \textsc{left}     & \phantom{0}5 & \phantom{0}5.0\,m & \textsc{stop} 2, \textsc{right} 2 \\
\bottomrule
\end{tabular}
\end{table}

\textbf{Neither grounded model grounds its lateral decisions.} Under reflection
across the lane axis the mirror swap rate is $0.000$ for both: $0/50$ directional
actions for Qwen and $0/13$ for Cosmos3. On Qwen's $50$ directional frames the
reflected image yields \textsc{slow} $25$ times, \textsc{left} $24$ and
\textsc{stop} once; \textsc{right} never appears in any Qwen output, in any
condition, on any frame. For that model \textsc{left} is not a lateral decision
but a token for \emph{the road is clear, proceed around}, and its effective action
space is three-valued and ordered purely by distance. Cosmos3 does emit both
directions, yet its reflected responses are equally unrelated to the flip. No
other test in the battery predicts this; the mirror probe is what isolates it,
and it now does so on two models that pass every other grounding criterion.

\textbf{MiniCPM-V-4.5 detects image presence, not content.} It is perfectly
deterministic ($s=0.000$) and separates road-like images from non-images
completely (\textsc{slow} $142$ / \textsc{straight} $2$ on real frames;
\textsc{stop} $144$ on each of blank, noise and text-only). Within real frames it
is effectively constant: $p=0.370$ against the obstacle, $I(A;g)=0.028$ bits at
$p=0.125$.

\textbf{Kimi-K3 is self-inconsistent, and only the noise floor reveals it.} With
a $4096$-token budget and structured output it parses on $99.3\%$ of calls, emits
all five actions, and registers $\mathrm{IDI}=0.771$, Cram\'er's $V=0.302$ at
$p<10^{-4}$, and $I(A;g)=0.122$ bits at $p=0.030$. On every test the battery
inherited from prior work it looks image-dependent. It reproduces its own answer
on $21\%$ of \emph{identical} images, giving $s=0.792$ and
$\mathrm{IDI}_{\mathrm{adj}}=-0.021$: the model disagrees with itself more than
it disagrees with a blank frame. Its apparent mirror swap rate of $0.225$ over
$40$ directional actions is likewise inside its own noise. We classify it as a
random policy with respect to the visual input, and we would have classified it
as grounded without the repeat condition.

\textbf{Kimi-K2.6 is not scorable.} Even at $4096$ tokens it exhausts the budget
on $28.2\%$ of calls, and on the parsed remainder its gap relationship is
inverted: it brakes at a median $27.3$\,m and proceeds at $6.2$\,m, consistently
across both runs. With $s=0.410$ and $p=0.483$ against the obstacle we exclude it
from conclusions.

A caveat on both Kimi results. K3's randomness is conditional on the serving
configuration that made it parseable: under a $1024$-token budget it was
truncated on $60\%$ of calls, and under JSON-mode output at $4096$ tokens it
answers in a mean of $36$ tokens and at random. Neither configuration we tried
yields a usable controller; we do not claim the model is inherently random.

\textbf{Replication.} The complete hosted battery, five models, $5\,760$ calls, was executed twice under identical configuration, thirty minutes apart. All
five verdicts reproduced. For the two grounded models the agreement is close:
$\mathrm{IDI}$ $0.604$ against $0.604$ and $0.778$ against $0.778$; $I(A;g)$
$0.466$ against $0.408$ and $0.159$ against $0.198$; mirror swap $0.000$ in all
four cases, on $51/50$ and $9/13$ directional actions. MiniCPM-V-4.5 was
bit-identical. Kimi-K3's noise floor was $0.722$ and $0.792$, and its
$\mathrm{IDI}_{\mathrm{adj}}$ changed sign between runs ($0.160$, $-0.021$),
which is itself consistent with a random policy. Because hosted endpoints can
change without notice, we report the later run as primary and the earlier as its
replicate.

\begin{figure}[t]
\centering
\includegraphics[width=\columnwidth]{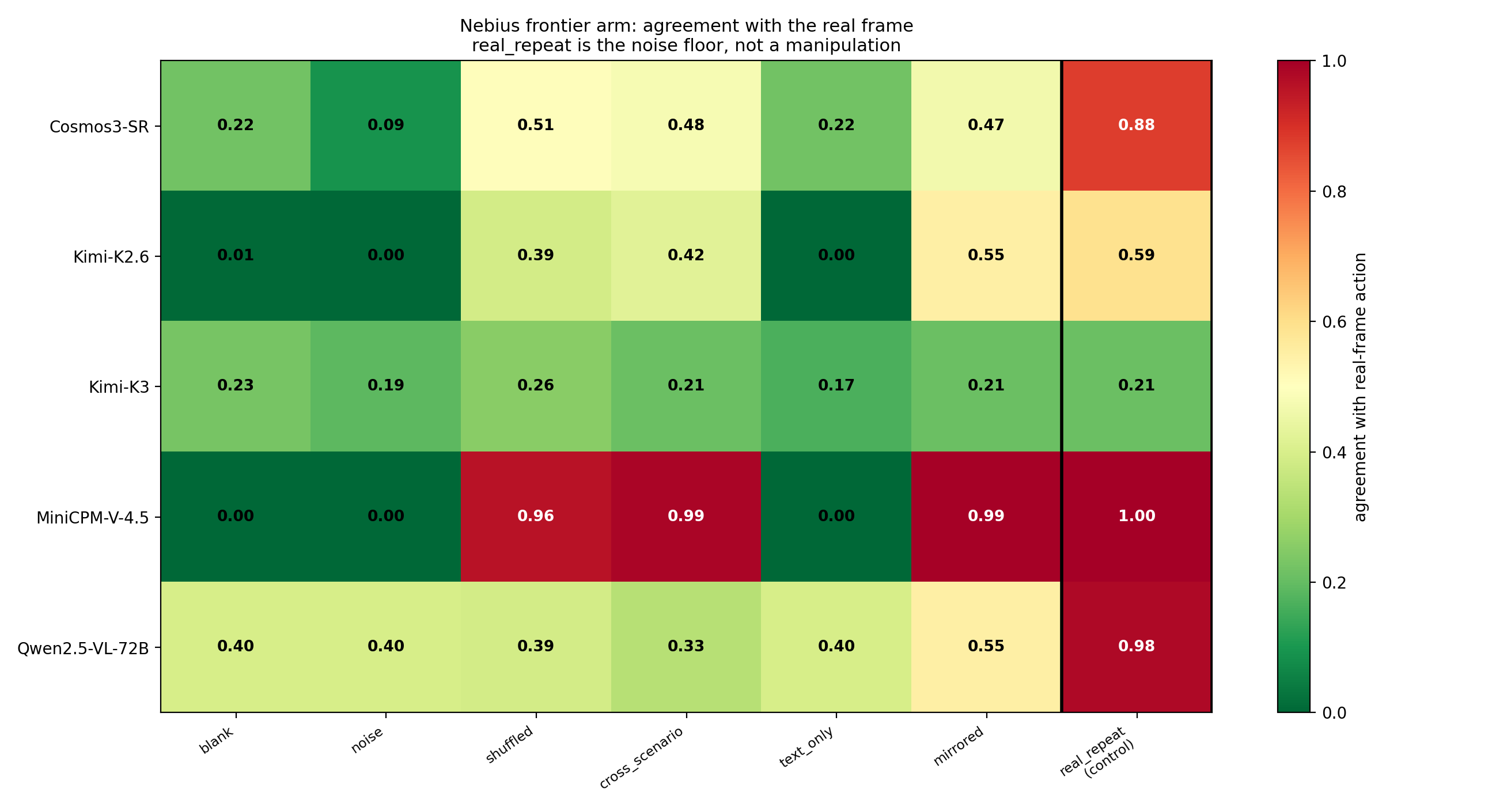}
\caption{Agreement with the real-frame action across blind-input conditions,
hosted arm. Red indicates the model produces the same action whether or not it
can see. \texttt{real\_repeat} is the noise floor, not a manipulation: a light
cell in that column means the model does not reproduce its own answer.}
\label{fgr:bicb}
\end{figure}

Figure~\ref{fgr:bicb} shows the full agreement matrix.

\subsection{Prompt formulation gates measured grounding}
\label{sn:prompts}

Table~\ref{t:prompts} ablates six prompt formulations. The effect is large and
partly counter-intuitive.

\begin{table}[t]
\caption{Prompt ablation, $100$ queries per cell. Modal share $1.00$ is a
constant policy; $n_a$ is the number of distinct actions emitted.}
\label{t:prompts}
\centering
\scriptsize
\begin{tabular}{@{}llrrr@{}}
\toprule
\textbf{Model} & \textbf{Prompt} & \textbf{Modal} & \textbf{$n_a$} & \textbf{IDI} \\
\midrule
\multirow{4}{*}{Qwen2.5-VL-72B}
 & baseline            & 0.37 & 3 & 0.63 \\
 & chain-of-thought    & 0.48 & \textbf{6} & 0.99 \\
 & few-shot            & 0.96 & 2 & 1.00 \\
 & anti-degeneracy     & 0.93 & 2 & 1.00 \\
\midrule
\multirow{4}{*}{MiniCPM-V-4.5}
 & baseline            & 0.98 & 2 & 1.00 \\
 & chain-of-thought    & \textbf{0.49} & \textbf{5} & 0.97 \\
 & few-shot            & 0.64 & 3 & 1.00 \\
 & anti-degeneracy     & 0.44 & 4 & 0.82 \\
\midrule
\multirow{4}{*}{SmolVLM}
 & baseline            & 0.89 & 2 & 0.11 \\
 & minimal             & \textbf{0.48} & 3 & 0.53 \\
 & few-shot            & \textbf{1.00} & 1 & 0.00 \\
 & anti-degeneracy     & 0.92 & 3 & 0.08 \\
\bottomrule
\end{tabular}
\end{table}

Chain-of-thought roughly halves modal share for MiniCPM-V-4.5 ($0.98
\rightarrow 0.49$) and more than doubles its action repertoire. A minimal prompt
does the same for SmolVLM ($0.89 \rightarrow 0.48$).

The counter-intuitive result concerns the two prompts designed to \emph{prevent}
degeneracy. Few-shot exemplars drive Qwen2.5-VL-72B from modal share $0.37$ to
$0.96$ and SmolVLM to a fully constant $1.00$. An instruction stating explicitly
that previous controllers had failed by emitting the same action on every frame,
and that different images must produce different actions, drives Qwen to $0.93$.
Prompts that constrain the output format appear to collapse the policy onto
whichever action the format makes most salient.

This is consistent with a body of work showing that language-model behaviour is
highly sensitive to prompt surface form
\cite{webson2022prompt,zhu2023promptbench,elazar2021measuring} and, specifically,
to the ordering and choice of few-shot exemplars \cite{lu2022fantastically}.

Two implications. First, any claim that a VLM cannot ground control must be
qualified by prompt formulation; a single-prompt null result is not evidence
about the model. Second, increased action entropy is not itself evidence of
grounding, Table~\ref{t:prompts} reports diversity, not scene-dependence,
and the two dissociate (Fig.~\ref{fgr:mi}). The same caution applies to
chain-of-thought \cite{wei2022chain}, whose verbalised reasoning is known not to
reliably reflect the computation that produced the answer
\cite{turpin2024language,lanham2023measuring}.

\subsection{Transfer and deployment characteristics}
\label{sn:cost}
Across eight model--platform cells, multilingual perception scores
\cite{deZarza2026} do not predict control grounding (Pearson $r=-0.438$,
$p=0.278$; $n=8$), so description scores remain proxies
\cite{belinkov2022probing}. Deployment is also non-monotone: the two grounded
hosted models are the cheapest tested ($\$0.078$ and $\$0.082$ per thousand
decisions), whereas unscorable Kimi-K2.6 costs $\$9.01$, averages $2\,186$
output tokens, truncates on 28\% of calls and has 11.3~s median latency. Complete
cost, latency and edge-energy tables are released.

\section{Discussion and Conclusion}

The central lesson of this study is that successful task completion and visual
grounding are not equivalent. A VLM controller can obtain a favourable reward,
travel a long distance, or avoid collisions without extracting the geometry
required for the task. We therefore separate three questions that are often
conflated: whether the model uses the image at all, whether it recovers
task-relevant spatial information, and whether that information supports an
appropriate control decision. Across $32\,874$ scored calls, the proposed
battery shows that a model may pass one of these levels while failing the next.
Some controllers react to the presence of an image but not to its content;
others distinguish near from distant obstacles but fail to exchange
\textsc{left} and \textsc{right} after reflection; and several obtain apparently
safe trajectories by repeatedly selecting a conservative action.

The direct-control experiments consequently provide a caution against
interpreting aggregate closed-loop metrics in isolation. A
constant-\textsc{slow} policy outperforms the scripted geometric controller on
the principal \texttt{highway-env} metrics, despite receiving no visual input.
Several edge models are nearly constant or image-invariant, while the hosted
models exhibit distinct regimes ranging from genuine longitudinal grounding to
image-presence sensitivity, stochastic self-inconsistency, and unscorable
truncation. In the structured local arm, no VLM passes the joint geometric gate:
the strongest models classify longitudinal proximity reasonably well, but their
distance estimates and lateral decisions remain inconsistent under lane-axis
reflection. The image-only positive control nevertheless recovers lead distance
to $0.090$~m MAE with exact mirror equivariance, demonstrating that the failure
does not arise because the stimuli or controller interface lack sufficient
information.

The results also identify a useful positive capability. When control is
decomposed rather than assigned monolithically to one model, the longitudinal
signal can be recovered reliably. The symmetry-consensus guardian combines
Gemma~4 and Qwen3.5 predictions from the original and reflected views and
achieves $0.954$ balanced accuracy on an untouched 272-frame holdout.
Furthermore, the same model pair and 2-of-4 threshold are recovered in all
12 nested leave-one-episode-out folds. Allowing the guardian to abstain when the
four votes split evenly raises committed balanced accuracy to $0.973$ at
$0.824$ coverage, while unanimous decisions reach $0.996$ at $0.647$ coverage.
This behaviour is important operationally: disagreement becomes a usable
uncertainty signal rather than an error that must be hidden behind a default
action. When deterministic perception retains responsibility for lateral
clearance, the resulting modular replay agrees with the renderer policy on
$0.934$ of holdout frames and preserves exact mirror equivariance.

The MPC experiments clarify the boundary of this modular strategy. A competent
low-level optimizer can accurately execute a valid reference and exploit known
obstacle geometry, but it cannot reconstruct geometry that is absent from the
high-level visual intent. Adding MPC downstream therefore improves tracking,
not grounding: collapsed or one-sided VLM directives remain collapsed after
optimization. The appropriate architectural conclusion is not to discard VLMs,
but to restrict their authority to capabilities that can be independently
validated. In the present setting, VLMs are most credible as selective
longitudinal hazard monitors, while metric geometry, constraint enforcement,
and actuation remain under deterministic or otherwise verifiable components.

More broadly, evaluations of VLM control should test each link of the
perception--decision--control chain. Blind inputs determine whether image
content matters; repeated identical inputs establish the model's stochastic
noise floor; metamorphic transformations test whether decisions obey known
geometric relations; and non-visual baselines reveal when the simulator rewards
a shortcut. These checks change the interpretation of the same trajectories
from ``the controller succeeds'' to the more precise statement of what the
model actually perceives and which decisions it can support. Current VLMs are
therefore not yet dependable general-purpose zero-shot controllers, but they can
already contribute useful semantic evidence within a bounded architecture based
on consensus, symmetry and abstention. Extending that authority will require
closing the lateral-grounding gap and validating the complete modular stack
under natural-camera observations, richer dynamics, and genuinely closed-loop
deployment conditions.

\section*{Code Availability}
All code, data, and experimental artifacts are publicly available at:
\url{https://github.com/drdecurto/VLControl}. The repository contains the
complete evaluation pipeline, including the input-ablation battery,
structured local controllers, MPC experiments, and the
symmetry-consensus guardian.

\section*{Acknowledgements}
This research was supported by the LUXEMBOURG Institute of
Science and Technology through the projects
``ADIALab-MAST'' and ``LLMs4EU''
(Grant Agreement No~101198470) and the BARCELONA
Supercomputing Center through the project ``TIFON''
(File number MIG-20232039). 

%% =====================================================================
% Balance the two reference columns on the final page.

\end{document}

%% file: tables/controller_local_vlms.tex
\begin{tabular}{@{}lrrrrrrr@{}}
\toprule
\textbf{Estimator} & \textbf{Parse} & \textbf{Bal.} & \textbf{MAE} & \textbf{Scene} & \textbf{Action} & \textbf{Swap ($n$)} & \textbf{Crash} \\
 &  & \textbf{acc.} & \textbf{(m)} & \textbf{eq.} & \textbf{eq.} &  &  \\
\midrule
\textbf{Renderer positive} & \textbf{1.000} & \textbf{1.000} & \textbf{0.090} & \textbf{1.000} & \textbf{1.000} & \textbf{1.000 (10)} & 0.167 \\
\midrule
Qwen3-VL-8B & 1.000 & 0.445 & 15.345 & 0.253 & 0.365 & 0.119 (134) & 0.083 \\
Gemma~4-12B & 1.000 & 0.858 & 16.014 & 0.399 & 0.497 & 0.000 (43) & 0.667 \\
Qwen3.5-9B & 1.000 & 0.867 & 13.125 & 0.031 & 0.538 & 0.097 (93) & 0.083 \\
Qwen2.5-VL-7B & 1.000 & 0.502 & 53.653 & 0.000 & 0.872 & 0.000 (1) & 1.000 \\
Ministral~3-8B & 1.000 & 0.534 & 6.240 & 0.010 & 0.556 & 0.134 (112) & 0.083 \\
MiniCPM-V-4.5 & 1.000 & 0.641 & 11.049 & 0.007 & 0.215 & 0.020 (199) & 0.083 \\
\bottomrule
\end{tabular}

%% file: tables/positive_guardian.tex
\begin{table}[t]
\centering
\caption{Leakage-controlled longitudinal guardian on the 272-frame holdout.}
\label{t:positive-guardian}
\scriptsize
\setlength{\tabcolsep}{1.8pt}
\resizebox{\columnwidth}{!}{%
\begin{tabular}{lrrrr}
\toprule
Method & Coverage & Acc. & Bal. acc. & Sens./Spec. \\
\midrule
Gemma4-12B, original & 1.000 & 0.842 & 0.869 & 0.821/0.917 \\
Two-model symmetry guardian & 1.000 & 0.956 & 0.954 & 0.958/0.950 \\
Guardian, abstain on 2--2 & 0.824 & 0.960 & 0.973 & 0.946/1.000 \\
Guardian, unanimous only & 0.647 & 0.994 & 0.996 & 0.992/1.000 \\
\bottomrule
\end{tabular}}
\end{table}

%% file: tables/mpc_intent_ablation.tex
\begin{tabular}{@{}lrrrr@{}}
\toprule
\textbf{Intent source} & \textbf{Goal} & \textbf{Crash} & \textbf{Dist. (m)} & \textbf{STOP} \\
\midrule
MPC-only & 23/25 & 0/25 & 95.6 & 0.0\% \\
oracle intent & 23/25 & 0/25 & 96.6 & 7.2\% \\
random intent & 14/25 & 0/25 & 82.6 & 21.4\% \\
Gemma4 VLM & 8/25 & 0/25 & 46.3 & 81.6\% \\
VLM, mirrored & 8/25 & 0/25 & 47.1 & 81.2\% \\
VLM, blank & 0/25 & 0/25 & 0.2 & 100.0\% \\
\bottomrule
\end{tabular}